\pdfoutput=1

\documentclass[11pt]{article}

\usepackage[]{emnlp2021}

\usepackage{times}
\usepackage{latexsym}
\usepackage{multirow}
\usepackage{amsmath}
\usepackage{courier}
\usepackage{graphicx}
\usepackage{enumitem}
\usepackage{tablefootnote}
\usepackage{pifont}
\usepackage{amsfonts}

\usepackage[T1]{fontenc}

\usepackage[utf8]{inputenc}

\usepackage{microtype}

\title{Connect-the-Dots: Bridging Semantics between Words and Definitions via Aligning Word Sense Inventories}

\author{Wenlin Yao \hspace{.04in} Xiaoman Pan \hspace{.04in} Lifeng Jin \hspace{.04in} Jianshu Chen \hspace{.04in} Dian Yu \hspace{.04in} Dong Yu\\
Tencent AI Lab, Bellevue, WA, USA\\
         {\tt \fontsize{10}{12}\selectfont \{wenlinyao,xiaomanpan,lifengjin,jianshuchen,yudian,dyu\}@tencent.com}\\ 
} 

\begin{document}
\maketitle
\begin{abstract}
Word Sense Disambiguation (WSD) aims to automatically identify the exact meaning of one word according to its context. 
Existing supervised models struggle to make correct predictions on rare word senses due to limited training data and can only select the best definition sentence 
from one predefined word sense inventory (e.g., WordNet). To address the data sparsity problem and generalize the model to be independent of one predefined inventory, we 
propose a gloss alignment algorithm that can align definition sentences (glosses) with the same meaning from different sense inventories to collect rich lexical knowledge.
We then train a model to identify semantic equivalence between 
a target word in context and one of its glosses
using these aligned inventories, which exhibits strong transfer capability to many WSD tasks\footnote{Models and code are available at \url{https://github.com/wenlinyao/EMNLP21-ConnectTheDots}. We will also release the checkpoint of the pretrained model for reproducibility.}.
Experiments on benchmark datasets show that the proposed method improves predictions on both frequent and rare word senses, outperforming prior work by 1.2\% on the All-Words WSD Task and 4.3\% on the Low-Shot WSD Task.
Evaluation on WiC Task also indicates that our method can better capture word meanings in context.
\end{abstract}

\section{Introduction}

\begin{figure*}[t]
 \centering
  \includegraphics[width = 6.3in]{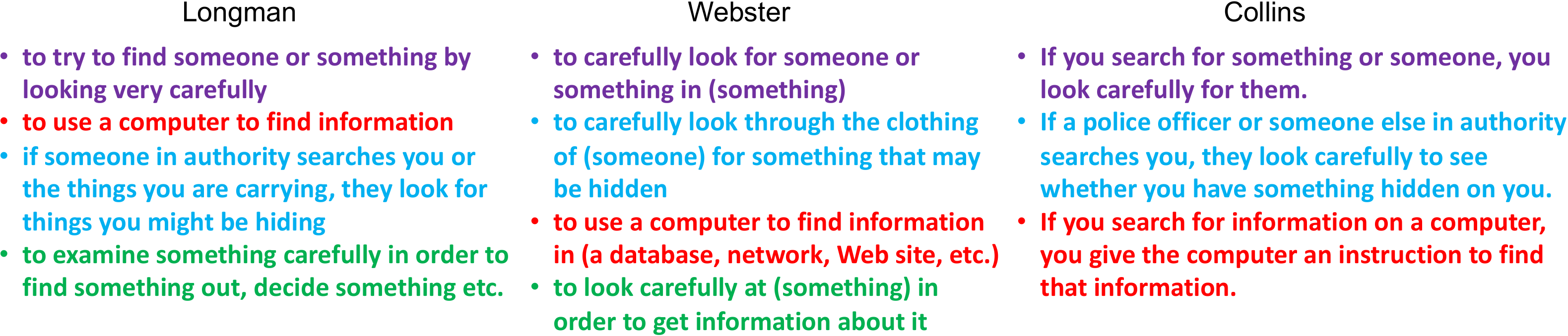}
 \caption{Definition sentences of word \textit{search} retrieved from three dictionaries: Longman Dictionary of Contemporary English, Merriam-Webster's Advanced Learner's Dictionary, and Collins  COBUILD Advanced Dictionary.}
\label{dict_def_examples}
\end{figure*}

Human language is inherently ambiguous since words can have various meanings in different contexts. Word Sense Disambiguation (WSD) aims to automatically identify the correct sense (meaning) of the target word within a context sentence, 
which is essential to many downstream tasks such as machine translation and information extraction.
Recently, many approaches have achieved state-of-the-art performance on WSD 
by fine-tuning language models pretrained with massive text data on task-specific datasets \citep{blevins2020moving,yap2020adapting}.

However, fine-tuning a WSD model using task-specific resources could limit its applicability and may cause two major problems.
First, the performance of models decreases significantly when predicting on rare and zero-shot word senses \citep{kumar2019zero,choubey2020one,blevins-etal-2021-fews} because there are no sufficient supporting examples in training data.
Second, the trained models are often inventory-dependent which can only select the best definition from one predefined word sense inventory (mainly WordNet) that human annotations are based upon.

In this paper, we overcome these limitations by leveraging abundant lexical knowledge from various word sense inventories. 
As we know, dictionaries that are compiled by experts contain rich sense knowledge of words. Moreover, a dictionary usually provides several example sentences for each word sense to illustrate its usage, which can be viewed as context sentences of that word sense.   
Since a word's sense (meaning) can be determined by its context, the word itself in a given context and the definition sentence corresponding to the correct sense are merely two surrogates of the same meaning (semantically equivalent). 
Furthermore, we observe that different dictionaries normally summarize meanings of a word to a close number of word senses, where definition sentences (glosses) from different dictionaries are different expressions of the same bunch of meanings. For example, Figure \ref{dict_def_examples} lists glosses retrieved from three dictionaries for verb word \textit{search}. We can see that glosses with the same color have the same meaning and can be aligned across different dictionaries. 

Based on this observation, we propose a gloss alignment algorithm to leverage abundant lexical knowledge from various word sense inventories. We convert the problem of aligning two groups of glosses according to meanings to an optimization problem -- Maximum Weighted Graph Matching -- to find the best matching that maximizes the overall textual similarity. 
In this way, we can gather general semantic equivalence knowledge from various dictionaries as a whole for all word senses, especially for rare senses that are less frequently 
seen in human-annotated data.

To make use of the derived semantic equivalence knowledge, we adopt a transfer learning approach that first pretrains a general semantic equivalence recognizer by contrasting the word representations in example sentences with the sentence representations of positive glosses or negative glosses. 
The general model can be directly applied to downstream WSD tasks or further fine-tuned on the task-specific dataset to get an expert model.
We test our two-stage transfer learning scheme on two WSD benchmark tasks, i.e., 
the standard task \citep{raganato2017word} that focuses on all-words WSD 
and FEWS \citep{blevins-etal-2021-fews} task that emphasizes low-shot (including few-shot and zero-shot) WSD. Experimental results show that the general model 
(without fine-tuning)
surpasses the supervised baseline by 13.1\% on zero-shot word senses.
After further fine-tuning with build-in training data, the expert model outperforms the previous state-of-the-art model 
by 1.2\% on all-words WSD tasks and 4.3\% on low-shot WSD tasks. 
Adding semantic equivalence knowledge to the Word-in-Context (WiC) task \citep{pilehvar2019wic} also improves the accuracy of RoBERTa\textsubscript{Large} \citep{liu2019roberta} by 6\%, which even outperforms the 9X larger T5 model \citep{raffel2020exploring}.

Overall, the major contributions of our work are two-fold. 1) We propose a gloss alignment algorithm that can integrate lexical knowledge from different word sense inventories to train a general semantic equivalence recognizer. 2) Without using task-specific training data, the general model not only performs well on overall word senses but demonstrates strong applicability to low-shot senses. The general model can turn into an expert model to achieve new state-of-the-art performance after further fine-tuning.



\section{Related Work}
\noindent\textbf{Supervised WSD Approaches.}
Most existing WSD models are learned in a supervised manner and depend on human-annotated data.
For example, 
\citet{raganato2017neural} regarded WSD as a sequence labeling task and trained a BiLSTM model with self-attention using multiple auxiliary losses.
\citet{luo2018leveraging} introduced a hierarchical co-attention mechanism to generate gloss and context representations that can attend to each other.
More recently, several BERT-based models have achieved new state-of-the-art performance on WSD by fine-tuning a pretrained language model. 
GlossBERT \citep{huang2019glossbert} appends each gloss to a given context sentence to create pseudo sentences and 
predicts them
as either positive or negative depending on whether the sense corresponds to the correct sense 
or not.
Bi-Encoder Model (BEM) \citep{blevins2020moving} represents the target words and senses in the same embedding space using a context encoder and a gloss encoder but optimizes on each word individually. 
\citet{yap2020adapting} formulated WSD as a relevance ranking task and fine-tuned BERT to select the most probable sense definition from 
candidate senses.
The neural architecture of our semantic equivalence recognizer realizes the benefits of GlossBERT and BEM.

\begin{figure*}[t]
 \centering
  \includegraphics[width = 6.3in]{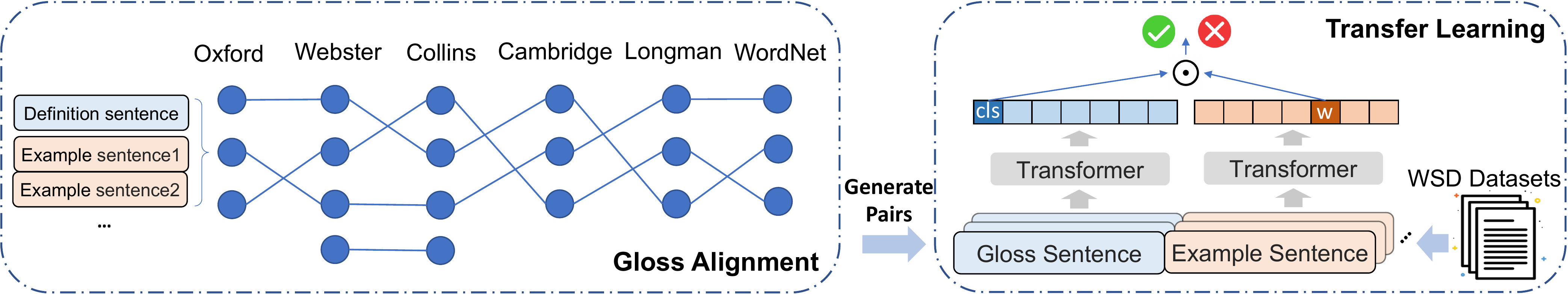}
 \caption{Overview of our approach. The left part illustrates the gloss alignment algorithm where each blue circle is a gloss containing one definition sentence and several example sentences. The right part is our model architecture to predict the  semantic equivalence of a word in context and a gloss by comparing their representations obtained from a shared transformer encoder.
 Task-specific WSD datasets can be further used to fine-tune our model.}
\label{model_pipeline}
\end{figure*}

\noindent\textbf{Knowledge-Based WSD Approaches.}
Closely related to our work, many knowledge-based approaches rely on Lexical Knowledge Bases (LKB), such as Wikipedia and WordNet, to enhance representations of word senses.
BabelNet \citep{navigli2010babelnet} creates a resource by automatically mapping 
encyclopedic knowledge (Wikipedia) to lexicographic knowledge (WordNet) with the aid of Machine Translation. 
Lesk \citep{basile2014enhanced} relies on a word-level similarity function to measure the semantic overlap between the context of a word and each sense definition. 
SENSEMBERT \citep{scarlini2020sensembert} produces high-quality latent semantic representations of word meanings 
by incorporating knowledge contained in BabelNet into language models.
Other approaches try to learn better gloss embeddings by considering the WordNet graph structure (e.g., hypernyms, hyponyms, synonyms, etc.) \citep{luo2018incorporating, loureiro2019language, kumar2019zero, bevilacqua2020breaking}. 
For example, \citet{kumar2019zero} proposed 
EWISE
to improve model's performance on rare or unseen senses by learning knowledge graph embeddings from WordNet. 
Building upon EWISE, \citet{bevilacqua2020breaking} developed a hybrid 
approach that incorporates more lexical knowledge (e.g., hypernymy, meronymy, similarity in WordNet) into 
the model through synset graph embeddings. 

\section{Overview of Our Approach}

Figure \ref{model_pipeline} shows the overview of our approach. 
We first collect all word glosses and corresponding example sentences from six word sense inventories. We next apply the gloss alignment algorithm to find the best matching between two groups of glosses retrieved from two different inventories for every common keyword. 
By contrasting example sentences with the correct glosses and incorrect glosses within each inventory or across different inventories, we automatically gather rich supervision for pretraining a universal binary classifier that can determine whether the keyword in the context sentence (example sentence) and a gloss are semantically equivalent or not.
The pretrained general model can be directly used in downstream WSD tasks or further fine-tuned to get an expert model.

\section{Aligning Glosses across Word Sense Inventories}

\subsection{Data Collection}

\begin{table}[t]
\small
\begin{center}
\setlength{\tabcolsep}{4.5pt}
\begin{tabular}{l|ccc|cc}
Inventory & Words & Glosses & ES & Gls/W & ES/W \\\hline
Oxford& 52.5K & 86.2K & 96.8K & 1.6 & 1.8  \\
Webster& 39.8K & 72.5K & 100.6K & 1.8 & 2.5  \\
Collins& 34.4K & 61.4K & 89.5K & 1.8 & 2.6 \\
Cambridge& 36.6K & 67.0K & 64.9K & 1.8 & 1.8 \\
Longman& 36.9K & 63.8K & 70.2K & 1.7 & 1.9  \\
WordNet& 147.3K & 206.9K & 47.4K & 1.4 & 0.3 \\\hline
\end{tabular}
\end{center}
\caption{Statistics of six word sense inventories used (phrases are included in word counting). ES: Example Sentences; Gls/W: average glosses per word; ES/W: average example sentences per word.}
\label{dict_statistics}
\end{table}

We collected word sense inventory data by querying WordNet 3.0 \citep{miller1995wordnet} and the electronic edition of five professional dictionaries for advanced English learners:
Oxford Advanced Learner's Dictionary
\citep{oxfordDict},
Merriam-Webster's Advanced Learner's Dictionary
\citep{websterDict},
Collins COBUILD Advanced Dictionary
\citep{collinsDict},
Cambridge Advanced Learner's Dictionary
\citep{cambridgeDict},
and Longman Dictionary of Contemporary English
\citep{longmanDict}. Advanced learners' dictionaries have a good characteristic that they usually provide abundant example sentences to illustrate the usage of different word senses in context, making it possible to generate strong supervision for training a classifier. Table \ref{dict_statistics} shows statistics of six word sense inventories used. In total, we collected 557.8K glosses and 469.4K example sentences.

\subsection{Gloss Alignment as a Maximum-weight Matching Problem}
Each word sense inventory is a lexical knowledge bank that provides example sentences for illustrating word senses, including senses less frequently seen in the real world. Moreover, we observe that different inventories usually provide parallel explanations of meanings for a given word (Figure \ref{dict_def_examples}).
Thus, if we can align explanations (glosses) from different inventories 
according to meanings, we can significantly expand lexical knowledge acquired, especially for rare word senses. Essentially, finding the best alignment between two groups of glosses can be converted to Maximum-weight Bipartite Matching Problem \citep{cormen2009introduction,duan2014linear} that aims to find a matching in a weighted bipartite graph that maximizes the sum of weights of the edges.

\subsection{Problem Formulation}
Given a keyword, suppose we retrieved two word sense sets $S_1$ and $S_2$ from two inventories, where each set consists of a list of definition sentences (glosses). Given a reward function $r$: $S_1\times S_2 \rightarrow \mathbb{R}$, we want to find a 
matching\footnote{Note that unbalanced matching (i.e., $S_1$ and $S_2$ are different in size) can be reduced to balanced matching by adding new vertices to the smaller part and assigning weight 0 to edges pointing to them.} $f: S_1 \rightarrow S_2$ such that the total rewards $\sum_{a\in S_1, f(a)\in S_2}r(a, f(a))$
is maximized. By finding the matching $f$, we will know the best alignment between two word sense sets $S_1$ and $S_2$. In this paper, we use the sentence-level textual similarity as the reward function to find the best word sense alignment.
To measure the textual similarity between two definition sentences, we apply a pretrained model SBERT \citep{reimers2019sentence} that has achieved state-of-the-art performance on many Semantic Textual Similarity (STS) tasks and Paraphrase Detection tasks. 
Specifically, we apply SBERT to $S_1$ and $S_2$ to get sentence embeddings and then calculate cosine similarity as the reward function.

\subsection{Solving Bipartite Graph Matching by Linear Programming}
The Maximum-weight Graph Matching problem can be solved by Linear Programming \citep{matousek2007understanding,cormen2009introduction}. For simplicity, let weight $w_{ij}$ denotes the textural similarity score between the i\textsuperscript{th} definition sentence in $S_1$ and the j\textsuperscript{th} definition sentence in $S_2$. We next introduce another variable $x_{ij}$ for each edge $(i, j)$. $x_{ij} = 1$ if the edge between $i$ and $j$ is contained in the matching and $x_{ij} = 0$ otherwise. The total weight of the matching is $\sum_{(i,j)\in S_1 \times S_2}w_{ij}x_{ij}$. To reflect every vertex is in exactly one edge in the matching, we add constraints $\sum_{j\in S_2}x_{ij}=1$ for $i\in S_1$, and $\sum_{i\in S_1}x_{ij}=1$ for $j\in S_2$, to guarantee that the variable $x$ represents a perfect matching. Our goal is to find a maximum-weight perfect matching such that above constraints are satisfied. To sum up, aligning glosses between two word sense inventories is equivalent to solving the following linear integer programming problem:
\vspace{-0.3cm}
\begin{align*}
\max\limits_{\{x_{ij}\}} \quad
\sum_{(i,j)\in S_1 \times S_2}w_{ij}&x_{ij} \\
\text{s.t.} \quad
\sum_{j\in S_2}x_{ij}=1,\ &i\in S_1\\
\sum_{i\in S_1}x_{ij}=1,\ &j\in S_2\\
x_{ij}\in \{0,1\},\ &i\in S_1, j\in S_2
\end{align*}

\vspace{-0.3cm}
In our implementation, we consider all possible inventory combinations (select two from six) and apply the gloss alignment solver\footnote{Our implementation is based on Scipy library (\url{https://www.scipy.org/}).} to all common words shared by two inventories. For each word, the gloss alignment solver is only applied to glosses under the same POS category. Overall, we obtain 704K gloss alignment links.

\subsection{Positive and Negative Training Instances}\label{pos_neg_pairs}

For a given word, the gloss alignment algorithm provides us the linking from word sense set $S_1$ in one inventory to $S_2$ in another inventory. Two glosses (e.g., $g\in S_1$ and $g'\in S_2$) have the same meaning if they are aligned by the algorithm or have a different meaning if they are not aligned. So we can pair 
the definition sentence of $g$ ($g'$) to each example sentence in $g'$ ($g$) to generate gloss-context pairs for training the semantic equivalence recognizer. 
Pairs are labeled as positive if $g$ and $g'$ are aligned or negative otherwise\footnote{If $S_1$ and $S_2$ have a different number of glosses for a given word, we ignore the extra glosses that are not aligned.}. In experiments, we 
only consider aligned gloss pairs with textual similarities above a threshold (see Section \ref{humanEval}) to further improve the quality of supervision. In total, we generate 421K positive and 538K negative gloss-context pairs across different inventories.

Pairs are also generated by contrasting glosses within each inventory individually. In detail, for every word in an inventory, 
we pair the gloss sentence with its example sentences to get positive gloss-context pairs or pair the gloss sentence with example sentences from another gloss within the inventory to get negative gloss-context pairs\footnote{We only contrast to glosses having the same POS tag to get negative instances.}. We generate 1.3M positive and 418K negative gloss-context pairs in this way.
Similarly, we also generate context-context pairs by contrasting example sentences in two glosses to reflect the task setting of WiC (Section \ref{WiC_task}).
\section{A Unified Neural Model for Recognizing Semantic Equivalence}

\subsection{Model Architecture}
This section introduces our model architecture (the right part of Figure \ref{model_pipeline}) for recognizing semantic equivalence. 
Inspired by \citet{blevins2020moving}, our model first uses an encoder to get the semantic representation of the target word (within its context sentence) or the gloss sentence.
Next, by comparing two representations, our model predicts whether they are semantically equivalent or not. 

\noindent\textbf{Semantic Encoder.} 
We apply a pretrained BERT model to get the contextual word representation of the target word (with its context) or the sentence representation of the gloss sentence. 
Specifically, given an input sentence $S$ padded by the start symbol \texttt{[CLS]} and the end symbol \texttt{[SEP]}, we first obtain $N$ contextualized embeddings $\{o_i\}^{N}_{i=1}$ for all tokens $\{t_i\}^{N}_{i=1}$ using BERT. We next select the contextualized embedding at the target word position\footnote{If the target word is a phrase or the target word is tokenized into multiple subword pieces by the tokenizer, we average all subword embeddings to get its representation.} when $S$ is a context sentence, or select the first output embedding $o_0$ (corresponding to the special token \texttt{[CLS]}) as the sentence representation when $S$ is a gloss sentence. 

\noindent\textbf{Learning Objective.} After deriving 
embeddings using BERT, 
both representations $u$ and $v$, together with element-wise difference $|u-v|$ and element-wise multiplication $u\cdot v$ are concatenated and 
multiplied with the trained weight $W_t\in \mathbb{R}^{4n\times 2}$ 
with a softmax prediction layer for binary classification (semantically equivalent or not):
\vspace{-0.2cm}
\begin{equation}\notag
p = \mathrm{softmax}(W_t[u, v, |u-v|, u\cdot v])
\vspace{-0.2cm}
\end{equation}
where $n$ is the dimension of the embeddings.
Our experiments consider two model sizes: 
\textbf{SemEq-Base} that is initialized with the pretrained BERT\textsubscript{Base} \citep{devlin2019bert} model
with 12 transformer block layers, 768 hidden size, 12 self-attention heads and \textbf{SemEq-Large} that is initialized with the pretrained RoBERTa\textsubscript{Large} \citep{liu2019roberta} model with 24 transformer block layers, 1024 hidden size, 16 self-attention heads\footnote{Our implementation was based on \url{https://github.com/huggingface/transformers}.}.
We train our model using binary cross-entropy loss and AdamW \citep{loshchilov2018decoupled} optimizer with initial learning rate \{1e-5, 5e-6, 2e-6\}, 0.2 dropout, batch size 64 and 10 training epochs.

\begin{table}[t]
\small
\begin{center}
\setlength{\tabcolsep}{4pt}
\begin{tabular}{l|ccccc}
           & Noun & Verb & Adj & Adv & ALL \\\hline
Percentage & 55.6\% & 20.6\%  & 20.2\% & 2.5\% & 100\% \\
Accuracy   & 0.90  &  0.81  & 0.88 & 0.85 & 0.87 \\\hline
\end{tabular}
\end{center}
\caption{Accuracy of the Gloss Alignment Algorithm.}
\label{alignment_acc}
\end{table}

\section{Evaluation}

\begin{table*}[t]
\small
\begin{center}
\setlength{\tabcolsep}{4.5pt}
\begin{tabular}{ll|cccc|c|cccc|cccc|c}
\hline
  & & \multicolumn{4}{c|}{Model difference}   & \multicolumn{1}{c|}{Dev} & \multicolumn{4}{c|}{Test} & \multicolumn{5}{c}{Concatenation of all Datasets} \\
& Models & TS & IK & GS & MS & SE07                    & SE2  & SE3 & SE13 & SE15 & Noun     & Verb     & Adj     & Adv     & ALL     \\ \hline
1& Most Frequent Sense      &\ding{51}&-&-&-&  54.5  & 65.6 & 66.0& 63.8 & 67.1 & 67.7     & 49.8     & 73.1    & 80.5    & 65.5   \\
2&Lesk\textsubscript{emb} \citeyearpar{basile2014enhanced}  &\ding{51}&-&\ding{51}&-& 56.7&  63.0  & 63.7 & 66.2& 64.6 & 70.0 & 51.1     & 51.7     & 80.6    & 64.2   \\
3&BiLSTM \citeyearpar{raganato2017neural}    &\ding{51}&-&-&-&  -     & 71.1 & 68.4& 64.8 & 68.3 & 69.5     & 55.9     & 76.2    & 82.4    & 68.4   \\
4&HCAN \citeyearpar{luo2018leveraging}       &\ding{51}&-&-&-&  -     & 72.8 & 70.3& 68.5 & 72.8 & 72.7     & 58.2     & 77.4    & 84.1    & 71.1   \\
5&EWISE \citeyearpar{kumar2019zero}  &\ding{51}&-&\ding{51}&-&  67.3     & 73.8 & 71.1& 69.4 & 74.5 & 74.0     & 60.2     & 78.0    & 82.1    & 71.8   \\
6&LMMS\textsubscript{BERT} \citeyearpar{loureiro2019language} &\ding{51}&-&\ding{51}&L&  68.1  & 76.3 & 75.6& 75.1 & 77.0 & -        & -        & -       & -       & 75.4   \\
7&GlossBERT \citeyearpar{huang2019glossbert}   &\ding{51}&-&-&B&  72.5  & 77.7 & 75.2& 76.1 & 80.4 & 79.3     & 66.9     & 78.2    & 86.4    & 77.0   \\
8&BEM  \citeyearpar{blevins2020moving}    &\ding{51}&-&-&B&  74.5  & 79.4 & 77.4& 79.7 & 81.7 & 81.4     & 68.5     & 83.0    & 87.9    & 79.0   \\
9&AdaptBERT\textsubscript{Large} \citeyearpar{yap2020adapting} &\ding{51}&S&-&L&  72.7  & 79.8 & 77.8& 79.7 & \textbf{84.4} & 82.6     & 68.5     & 82.1    & 86.4    & 79.5   \\
10&EWISER \citeyearpar{bevilacqua2020breaking}&\ding{51}&S&\ding{51}&L&  \textbf{75.2}  & 80.8 & 79.0& 80.7 & 81.8 & 82.9     & 69.4     & \textbf{83.6}    & 87.3    & 80.1   \\
11&SemEq-Base &\ding{51}&-&-&B& 72.7	&79.0	&77.2	&78.0 &80.8		&81.0	&67.1	&81.7	&86.7 &78.2 \\
\hline
\multicolumn{11}{l}{\textbf{Ours: Data Augmentation}}\\ \hline
12&SemEq-Base  &\ding{51}&M&-&B& 73.2	&81.2	&77.7	&79.1 &81.5		&81.9	&68.9	&83.2	&87.6 &79.4 \\\hline
\multicolumn{11}{l}{\textbf{Ours: Transfer Learning}}\\ \hline
13&SemEq-Base-General &-&M&-&B& 65.7	&75.3	&70.9	&78.0	&79.8	&78.2	&61.3	&81.2	&80.3 &74.8 \\
14&SemEq-Base-Expert &\ding{51}&M&-&B&  74.1	&81.0	&{78.5}	&79.9	&82.6		&82.5	&{69.9}	&82.5	&\textbf{88.4} &79.9 \\
15&SemEq-Large-General &-&M&-&L&  65.1	& 76.1&74.3	&78.0	&83.0	& 79.1&	64.7& 82.3& 81.8 &76.4 \\
16&SemEq-Large-Expert &\ding{51}&M&-&L&  74.9	&\textbf{81.8} &\textbf{79.6}	& \textbf{81.2}&	81.8& \textbf{83.2} & \textbf{71.1} & 83.2 & 87.9 &\textbf{80.7} \\\hline
\end{tabular}
\end{center}
\caption{F1-score (\%) on All-Words WSD benchmark datasets. We distinguish models based on 1) using the Training Set (TS) SemCor or not, 2) using single (S) Inventory Knowledge (IK) (i.e., WordNet) or our multi-source (M) inventory knowledge, 3) using WordNet synset Graph Structures (GS) or not, and 4) transformer Model Size (MS) of Base (B) or Large (L).
Baseline systems are: Lesk\textsubscript{emb} \citep{basile2014enhanced}, Babelfy \citep{moro2015semeval}, BiLSTM \citep{raganato2017neural}, HCAN \citep{luo2018leveraging}, EWISE \citep{kumar2019zero}, LMMS\textsubscript{BERT} \citep{loureiro2019language}, GlossBERT \citep{huang2019glossbert}, BEM  \citep{blevins2020moving}, AdaptBERT\textsubscript{Large} \citep{yap2020adapting}, and EWISER \citep{bevilacqua2020breaking}.}
\label{WSD_all_perf}
\end{table*}

\subsection{Accuracy of the Gloss Alignment Algorithm}\label{humanEval}

To evaluate the accuracy of the gloss alignment algorithm, we randomly sample 1,000 gloss pairs from 704K alignments and ask two human annotators to judge whether two gloss sentences refer to the same meaning or not. Two annotators labeled 200 gloss pairs in common and agreed on 94\% (188) of them, achieving the kappa inter-agreement score of 0.74. One gloss pair is regarded as correct when both annotators label it as correct, and the remaining 800 gloss pairs were evenly allocated to two annotators to label. Table \ref{alignment_acc} shows the accuracy of the gloss alignment algorithm on each POS type based on human annotations. The accuracy on Noun, Verb, Adjective and Adverb words is 0.90, 0.81, 0.88 and 0.85, respectively, with an overall accuracy of 0.87.
In experiments, we apply a threshold of 0.6 to alignment results and only consider aligned gloss pairs with textual similarities above it, which can further improve gloss alignment accuracy to 0.98 based on human annotations. 
In this way, we can significantly improve the quality of training data that are generated from the automatically aligned dictionaries.

\subsection{Experiments on WSD}

We evaluate our model on two WSD datasets, i.e., WSD tasks standardized by \citet{raganato2017word} that focuses on all-words WSD evaluation and FEWS dataset proposed by \citet{blevins-etal-2021-fews} that emphasizes low-shot WSD evaluation.
Since both datasets are annotated using word senses in WordNet 3.0 \citep{miller1995wordnet}, we pair the context sentence 
with the annotated gloss in WordNet 3.0 to generate positive gloss-context instances or
other glosses of the word to get negative gloss-context instances for training. In validation or test, we apply the trained classifier to examine all possible glosses of the target word in WordNet 3.0 and select the gloss with the highest probability score as the prediction.
To incorporate rich lexical knowledge harvested from word sense inventories into model training, we consider two strategies:

\noindent\textbf{Data Augmentation.} We directly augment the build-in training set from each WSD dataset with gloss-context pairs generated from our aligned word sense inventories and then train the semantic equivalence recognizer (SemEq) to do WSD.

\noindent\textbf{Transfer Learning.} We first train our semantic equivalence recognizer ONLY using gloss-context pairs generated from our aligned word sense inventories. The trained classifier is a general model (\textbf{SemEq-General}) capable of deciding whether a gloss sentence and the target word in a context sentence are semantically equivalent independent from any specific word sense inventories. Next, to evaluate on a specific WSD dataset, we further fine-tune the general model on the build-in training set to get an expert model (\textbf{SemEq-Expert}). The expert model can adapt to the new domain to achieve better performance.

\begin{figure*}[t]
 \centering
  \includegraphics[width = 6.2in]{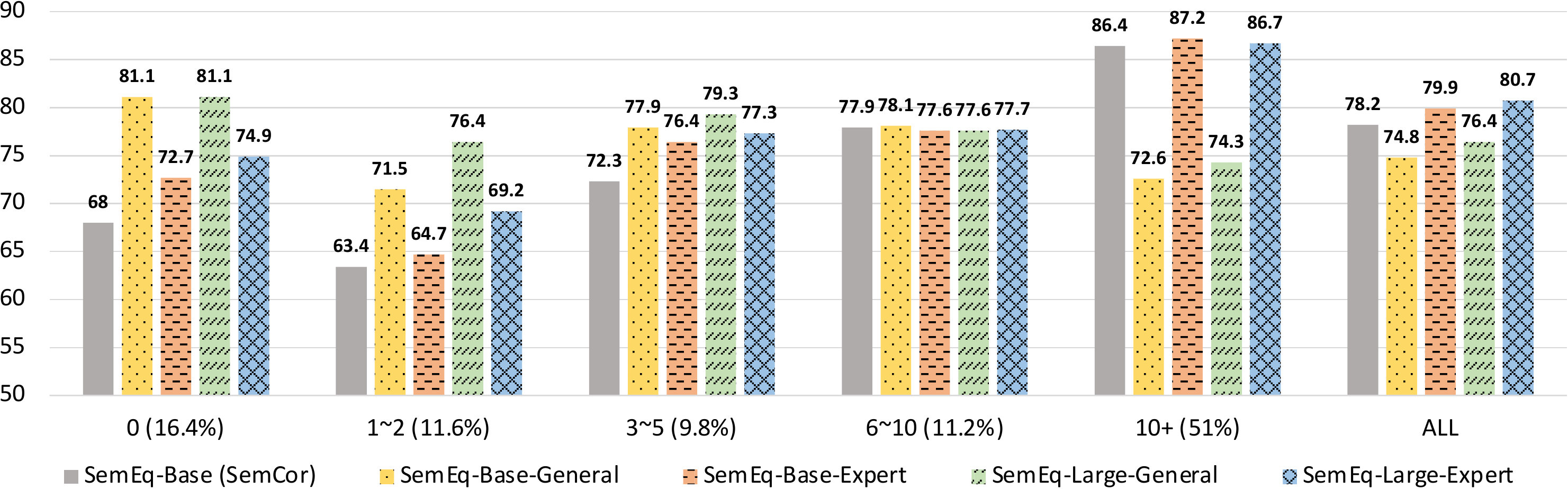}
 \caption{Evaluation (F1-score \%) on the aggregated ALL set of All-Words WSD when we separate word senses based on their training instance numbers in SemCor.}
\label{WSD_bar_chart}
\end{figure*}

\subsubsection{All-Words WSD Tasks}

We evaluate our model on the all-words WSD framework established by \citet{raganato2017word}. 
The testing dataset contains 5 benchmark datasets from previous Senseval and SemEval competitions, including Senseval-2 (SE2) \citep{edmonds2001senseval}, Senseval-3 (SE3) \citep{mihalcea2004senseval}, SemEval-07 (SE07) \citep{pradhan2007semeval}, SemEval-13 (SE13) \citep{navigli2013semeval}, and SemEval-15 (SE15) \citep{moro2015semeval}. Following \citet{raganato2017word} and other previous work, we use SemCor \citep{miller1993semantic} that contains 226,036 annotated instances as the build-in training set and choose SemEval-07 as the development set for hyper-parameter tuning. Since all datasets are mapped to word senses in WordNet 3.0 \citep{miller1995wordnet}, we retrieve all definition sentences of the target word from WordNet 3.0 to form gloss-context pairs for both training and testing.

Table \ref{WSD_all_perf} shows experimental results on all-words WSD datasets \citep{raganato2017word}. We also report models' performance on each POS category.
The first section includes results of the most frequent sense baseline and previous WSD models. 

The second section presents results of our model that adopt \textbf{data augmentation} strategy to incorporate multi-source inventory knowledge. 
SemEq-Base (line 11) is our model's performance when fine-tuning BERT\textsubscript{Base} sentence encoder only on the build-in SemCor training set. Compared to line 11, when augmenting SemCor with our multi-source inventory knowledge, the same model (line 12) improves the F1 on the aggregated ALL set by 1.2\%.

The third section of Table \ref{WSD_all_perf} reports the results of applying \textbf{transfer learning} strategy to exploiting our multi-source inventory knowledge. 
By only training on our multi-source inventory knowledge (without using SemCor), our model SemEq-Base-General (line 13) already achieves comparable performance with LMMS\textsubscript{BERT} (line 6, which is based on BERT\textsubscript{Large}). After further fine-tuning on the training set - Semcor, SemEq-Base-Expert (line 14) improves the performance on ALL to 79.9\%, which is slightly better than using the data augmentation strategy.
Moreover, increasing BERT model parameters 
(line 16)
further boosts the WSD performance on ALL to 80.7\%\footnote{We also tried BERT\textsubscript{Large} which is slightly worse than RoBERTa\textsubscript{Large}.}.

Overall, our SemEq-Large-Expert model (line 16) consistently outperforms AdaptBERT
\citep{yap2020adapting} (line 9), the previous best model without using WordNet synset graph information, 
on SE07, SE2, SE3 and SE13, attaining 1.2\% higher F1 on ALL.
The SemEq-Large-Expert model also better disambiguates all types of words including nouns, verbs, adjectives, and adverbs than AdaptBERT. It clearly demonstrates the benefits of leveraging multiple word sense inventories via automatic alignment and transfer learning. Our final model is 0.6\% higher even compared with EWISER \citep{bevilacqua2020breaking} that uses the \emph{extra} WordNet graph knowledge. 
We can see that by pretraining on lexical knowledge derived from aligned inventories, our model generalizes more easily and better captures semantic equivalence between the target word and a gloss sentence for identifying the correct word meaning.

\begin{table*}[t]
\small
\begin{center}
\setlength{\tabcolsep}{3pt}
\begin{tabular}{ll|c|ccc|ccc}
\hline
 & &     & \multicolumn{3}{c|}{Dev} & \multicolumn{3}{c}{Test} \\
&Models &TS &  Full Set & Few-shot & Zero-shot & Full Set  & Few-shot & Zero-shot     \\ \hline
1&Most Frequent Sense &\ding{51} & 26.4 & 52.8 & 0.0 &25.7 & 51.5 & 0.0 \\
2&Lesk\textsubscript{emb} \citep{basile2014enhanced} &\ding{51}& 42.5 & 44.9 & 40.1 &41.5 & 44.1 & 39.0  \\
3&BEM\citep{blevins2020moving} &\ding{51}& 73.8 & {79.3} & 68.3 &72.8 & {79.1} & 66.5 \\
4&BEM\textsubscript{SemCor} \citep{blevins-etal-2021-fews} &\ding{51}&  74.4 & 79.7 & 69.0 & 73.0 & 78.9 & 67.1 \\
5&SemEq-Base &\ding{51}& 73.5 & 78.7	&68.3	&72.4 &78.5	&66.3\\
\hline
\multicolumn{7}{l}{\textbf{Ours: Data Augmentation}}        \\\hline
6&SemEq-Base (+ WSI) &\ding{51}& 74.2 &78.4	&69.9	&73.7 &78.6	&68.7\\\hline
\multicolumn{7}{l}{\textbf{Ours: Transfer Learning}}        \\\hline
7&SemEq-Base-General &-&  68.2 &68.6	&67.8	&67.0 &67.7	&66.3\\
8&SemEq-Base-Expert &\ding{51}&  76.0 &{80.4}	&{71.5} &75.2	&{80.1}	&{70.2}\\
9&SemEq-Large-General &-&  70.7 &70.9	&70.5 &	69.8&71.2	&68.4\\
10&SemEq-Large-Expert &\ding{51}& \textbf{77.8} &\textbf{81.8}	&\textbf{73.7}	& \textbf{77.3}&\textbf{82.3}	&\textbf{72.2}\\
\hline
\end{tabular}
\end{center}
\caption{F1-score (\%) on the FEWS Low-Shot WSD benchmark dataset. 
WSI refers to knowledge extracted from aligned Word Sense Inventories. TS stands for the Training Set of FEWS.}
\label{WSD_lowshot_perf}
\end{table*}

\begin{table}[t]
\small
\begin{center}
\setlength{\tabcolsep}{4pt}
\begin{tabular}{lcc}\hline
Model & Acc.  & Parameters \\\hline
BERT\textsubscript{Large} \citep{devlin2019bert} & 69.6 & 340M \\
RoBERTa\textsubscript{Large} \citep{liu2019roberta} & 69.9 & 355M \\
KnowBERT\textsubscript{W+W} \citep{peters2019knowledge} & 70.9 & 523M \\
SenseBERT\textsubscript{Large} \citep{levine2020sensebert} & 72.1 & 380M \\
T5-Large \citep{raffel2020exploring} & 69.3 & 770M \\
T5-3B \citep{raffel2020exploring} & 72.1 & 3000M \\
BERT\textsubscript{ARES} \citep{scarlini2020more} & 72.2 & 342M \\\hline
SemEq-Large (+WSI) & \textbf{75.9} & 355M \\\hline
\end{tabular}
\end{center}
\caption{Accuracy (\%) on the WiC benchmark dataset.}
\label{WiC_perf}
\end{table}

In order to understand our model's behavior of transferring semantic equivalence knowledge from our word sense inventories to a specific WSD task, we partition word senses in the test set into groups according to their numbers of training instances found in the training set SemCor. 
As shown in Figure \ref{WSD_bar_chart}, 
by pretraining on our semantic equivalence knowledge and then fine-tuning on SemCor, SemEq-Base-Expert beats SemEq-Base (SemCor) that is only trained on SemCor across all annotation-rich and annotation-lacking word senses. Interestingly, without fine-tuning on SemCor, the general model (SemEq-Base-General) works surprisingly well on low-shot senses,
which is 13.1\%, 8.1\% and 5.6\% higher than SemEq-Base (SemCor) on 0 shot, 1-2 shot, 3-5 shot senses, respectively. After fine-tuning on SemCor, the expert models fit to the distribution of senses in the real world and achieve better overall performance.

\subsubsection{Few-Shot and Zero-Shot WSD Tasks}
By pretraining on massive semantic equivalence knowledge 
generated from aligned word sense inventories,
we expect our model performs better on annotation-lacking senses. We next evaluate our model on the FEWS dataset \citep{blevins-etal-2021-fews}, a new WSD dataset that focuses on low-shot WSD evaluation. FEWS is a comprehensive evaluation dataset constructed from Wiktionary and covers 35K polysemous words and 71K senses. Overall, the build-in training set of FEWS consists 87K sentence instances. The test (development) set consists of two evaluation subsets, i.e., a few-shot evaluation set and a zero-shot evaluation set; each subset contains 5K instances. Word senses that are used in zero-shot evaluation sets are verified to not occur in the training set, and word senses in few-shot evaluation sets will only occur 2 to 4 times in the training set.

Table \ref{WSD_lowshot_perf} presents the results on FEWS. BEM\textsubscript{SemCor} (line 4) is a similar transfer learning model but fine-tuned on SemCor before training on FEWS while BEM (line 3) only trains on FEWS. The second section of Table \ref{WSD_lowshot_perf} shows that augmenting the FEWS train set with our multi-source inventory knowledge (line 6) greatly improves zero-shot learning performance by 1.6\% on the dev set and 2.4\% on the test set (compared with line 5). Surprisingly, when we adopt the transfer learning strategy, the final SemEq-Large-Expert (line 10) model's performance on test sets increases to 82.3\% on few-shot senses and 72.2\% on zero-shot senses, which significantly outperforms all baseline models.

\subsection{Experiments on Context-Sensitive Word Meanings}\label{WiC_task}

Word-in-Context (WiC) Task \citep{pilehvar2019wic} from SuperGLUE benchmark \citep{wang2019superglue} provides a high-quality dataset for the evaluation of context-sensitive word meanings. WiC removes predefined word senses and reduces meaning identification to a binary classification problem in which, given two sentences containing the same lemma word, a model is asked to predict whether the two target words have the same meaning. Considering WiC uses WordNet as one lexical resource in its data construction,
we completely remove WordNet 
from our inventory knowledge 
to avoid data leaking.
Specifically, we simply add context-context pairs\footnote{We generate 3.3M positive pairs and 1.7M negative pairs.} generated from the other five inventories to the training set of WiC
to train a semantic equivalence recognizer.
Table \ref{WiC_perf} shows results on the WiC task comparing to other 
models\footnote{
We submit our model predictions to the competition page of WiC (\url{https://competitions.codalab.org/competitions
%/20010#participate
}) to get the test results.}. The results indicate that incorporating semantic equivalence knowledge from aligned inventories improves RoBERTa\textsubscript{Large}'s performance by 6\%, which also surpasses a large language model T5-3B (9X parameters) by 3.8\%. It demonstrates the superiority of incorporating our high-quality multi-source lexical knowledge
than blindly increasing the size of plain pretraining texts in language models.

\section{Conclusion}
Based on the observation that glosses of a word from different inventories usually are different expressions of a few meanings, we have proposed a gloss alignment algorithm that can
unify different lexical resources as a whole to generate abundant semantic equivalence knowledge.
The general model pretrained on derived equivalence knowledge can serve as a universal recognizer for word meanings in context or adapt to a specific WSD task by fine-tuning to achieve new state-of-the-art performance.
Our results also point to an interesting future research direction: how to develop a robust fine-tuning approach that is able to retain the excellent performance of the general model on low-resource senses while still improving performance on high-resource senses.

\section*{Ethical Considerations}
Copyrights of data used in this paper belong to their respective owners. The authors are permitted to use data under the permission of the non-commercial research purpose and following the principle of fair use. The authors will not reproduce, republish, distribute, transmit, or link data used on any other website without the express permission of respective owners. The authors bear the responsibility to comply with the rules of copyright holders.

\bibliography{anthology,custom}
\bibliographystyle{acl_natbib}


\end{document}